\documentclass[conference]{IEEEtran}
\usepackage{times}
\usepackage{graphicx}
\usepackage[numbers]{natbib}
\usepackage{multicol}
\usepackage{color}
\usepackage{epsfig}
\usepackage{graphicx}
\usepackage{algorithm,algorithmic}

\usepackage{adjustbox}
\usepackage{array}
\usepackage{booktabs}
\usepackage{colortbl}
\usepackage{float,wrapfig}
\usepackage{framed}
\usepackage{hhline}
\usepackage{multirow}
\usepackage{subcaption} 
\usepackage[font=small]{caption}
\usepackage[percent]{overpic}
\usepackage{siunitx}
\usepackage{amsmath,amsfonts,amssymb}

\DeclareMathOperator*{\argmin}{arg\,min}

\let\labelindent\relax
\usepackage{amsthm} 
\usepackage{bm}
\usepackage{nicefrac}
\usepackage{microtype}
\usepackage{contour}
\usepackage{courier}

\usepackage{changepage}
\usepackage{extramarks}
\usepackage{fancyhdr}
\usepackage{lastpage}
\usepackage{setspace}
\usepackage{soul}
\usepackage{xspace}
\usepackage{cuted}
\usepackage{fancybox}
\usepackage{afterpage}

\usepackage[pagebackref=true,breaklinks=true,colorlinks]{hyperref}
\usepackage{quoting}
\usepackage{epigraph}

\usepackage{paralist,tabularx}
\usepackage{comment}
\usepackage{pdfpages}

\usepackage{enumitem}

\makeatletter
\DeclareRobustCommand\onedot{\futurelet\@let@token\@onedot}
\def\@onedot{\ifx\@let@token.\else.\null\fi\xspace}

\makeatother

\definecolor{MyDarkBlue}{rgb}{0,0.08,1}
\definecolor{airforceblue}{rgb}{0.36, 0.54, 0.66}
\definecolor{MyDarkGreen}{rgb}{0.02,0.6,0.02}
\definecolor{MyDarkRed}{rgb}{0.8,0.02,0.02}
\definecolor{MyDarkOrange}{rgb}{0.40,0.2,0.02}
\definecolor{MyPurple}{RGB}{111,0,255}
\definecolor{MyRed}{rgb}{1.0,0.0,0.0}
\definecolor{MyGold}{rgb}{0.75,0.6,0.12}
\definecolor{MyDarkgray}{rgb}{0.66, 0.66, 0.66}

\def\namel{Iterative Residual Policy}
\def\names{IRP}

\begin{document}

\title{ Iterative Residual Policy \\ \LARGE{for Goal-Conditioned Dynamic Manipulation of Deformable Objects}}


\author{Author Names Omitted for Anonymous Review. Paper-ID: [3]} 
\author{Cheng Chi$^1$, Benjamin Burchfiel$^2$, Eric Cousineau$^2$, Siyuan Feng$^2$, Shuran Song$^1$ \\ $^1$ Columbia University \quad\quad\quad  $^2$ Toyota Research Institute
\\ \href{https://irp.cs.columbia.edu}{https://irp.cs.columbia.edu}}
\maketitle

\begin{abstract}
This paper tackles the task of goal-conditioned dynamic manipulation of deformable objects. This task is highly challenging due to its complex dynamics (introduced by object deformation and high-speed action) and strict task requirements (defined by a precise goal specification). To address these challenges, we present Iterative Residual Policy (IRP), a general learning framework applicable to repeatable tasks with complex dynamics. IRP learns an implicit policy via delta dynamics -- instead of modeling the entire dynamical system and inferring actions from that model, IRP learns delta dynamics that predict the effects of delta action on the previously-observed trajectory. When combined with adaptive action sampling, the system can quickly optimize its actions online to reach a specified goal. We demonstrate the effectiveness of IRP on two tasks: whipping a rope to hit a target point and swinging a cloth to reach a target pose. Despite being trained only in simulation on a fixed robot setup, IRP is able to efficiently generalize to noisy real-world dynamics, new objects with unseen physical properties, and even different robot hardware embodiments, demonstrating its excellent generalization capability relative to alternative approaches. 

\end{abstract}

\IEEEpeerreviewmaketitle

\section{Introduction}
\begin{figure}[t]
    \centering
    \includegraphics[width=\linewidth]{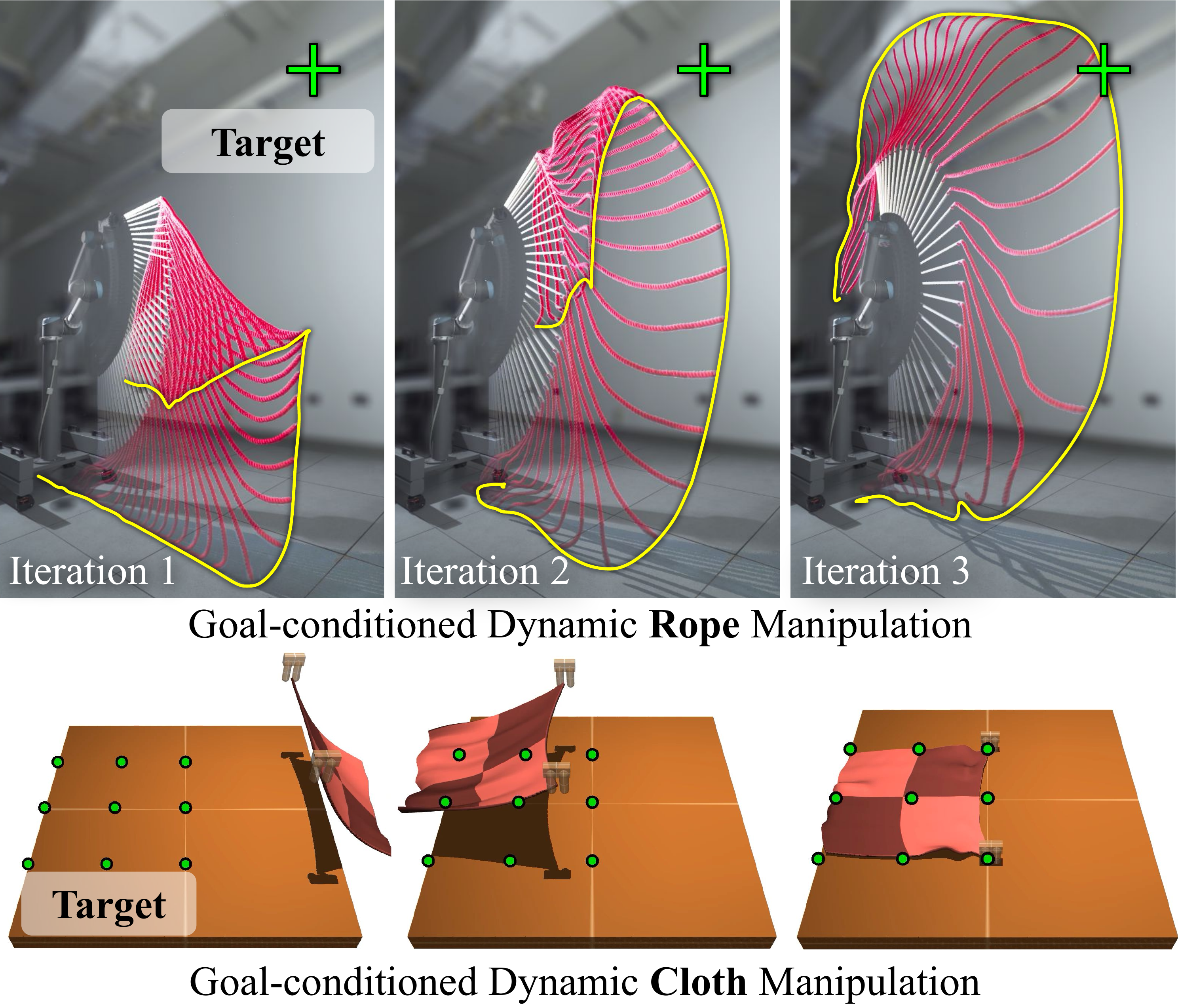}
    \caption{\textbf{Goal-conditioned dynamic manipulation of deformable objects.}  Our method achieves high accuracy by iteratively adjusting actions using visual feedback and a learned delta dynamics network. Top: Rope manipulation task. Goal configuration is defined by a target tip position (green cross).  Bottom: Cloth manipulation task. Goal configuration is defined by target keypoint locations (green dots). }
    \vspace{-5mm}
    \label{fig:teaser}
\end{figure}

We study the task of goal-conditioned dynamic manipulation of deformable objects, examples of which include whipping a target point with the tip of a rope or spreading a cloth into a target pose (Fig. \ref{fig:teaser}). 
Humans show remarkable ability to not only perform these tasks with high precision (e.g., aiming a lasso or setting a table cloth), but also to quickly transfer these manipulation skills to new objects with very few attempts. 
For robots, however, this has historically been a highly challenging problem due to several factors:

\begin{itemize}[leftmargin=3mm]
\item \textbf{Complex dynamics:} Unlike quasi-static manipulation, dynamic manipulations (often with high-speed actions) often lead to complex and non-linear effects. 
These dynamical processes are difficult to model, and pose significant challenges for system identification and state estimation.
As a result, it is often infeasible to apply classical algorithms, such as optimal control, when an accurate and performant forward model of the system is prohibitively difficult to obtain.

\item \textbf{Complex object properties:} 
Unlike rigid objects, deformable objects exhibit dynamics that are influenced by numerous hard-to-estimate factors such as nonlinear anisotropic stiffness, friction, density distribution, and changing aerodynamics. Not only is it challenging to estimate these properties, it is hard to map them to the parameter used in common simulators using simplified physics models.



\item \textbf{Precise goal conditions:}  Due to these challenges, prior work that applies dynamic manipulation to deformable objects generally studies tasks with low precision requirements such as cloth unfolding \cite{ha2021flingbot} or cable vaulting \cite{zhang2021robots}. In contrast, the tasks studied in this paper require precise actions in order to reach the specified goal conditions --- for example hitting a goal location with the tip of a whip or reaching a goal configurations defined by keypoints on a cloth.

\end{itemize}

To address these challenges, we present \textbf{\namel~(\names)}, a general formulation for goal-conditioned dynamic manipulation, that learns how to iteratively improve its actions to achieve a given goal condition using visual feedback.
This formulation highlights two key features: 

\begin{figure*}[t]
    \centering
    \includegraphics[width=\linewidth]{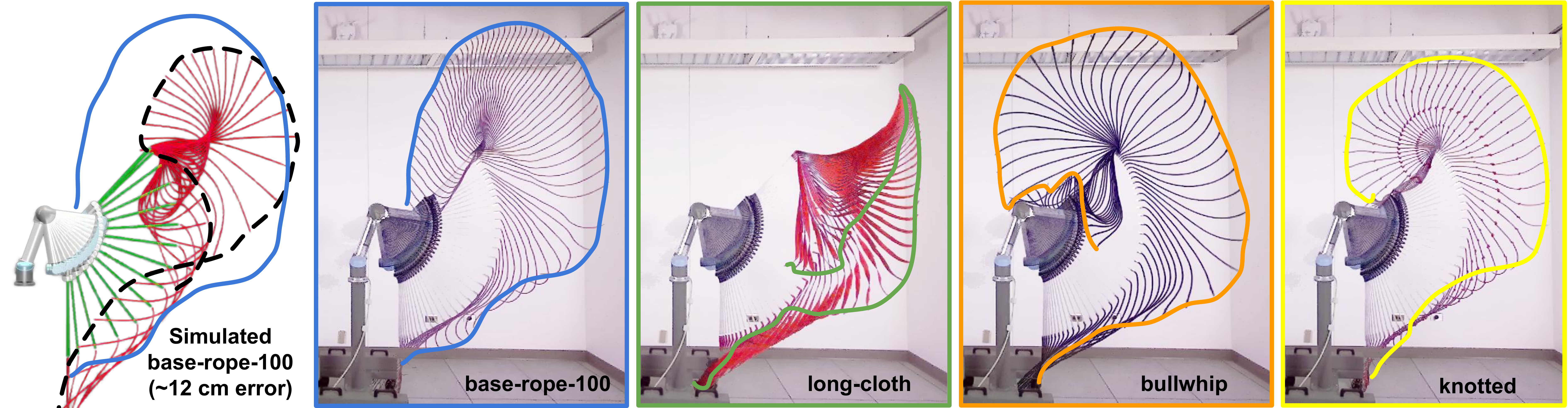}
    \caption{\textbf{Challenges.} Here we show the different rope trajectories under the \textbf{same} robot action. Due to different physical properties (e.g., aerodynamic, mass distribution or stiffness), the resulting trajectories varies drastically. Left plot shows the trajectory overlay between the real-world (solid curve) and simulated result (dashed curve) both for base-rope-100. The offset between them indicates the sim2real gap even with the accurate rope parameters.  These examples highlight the importance of online adaptation that is necessary for both sim2real transfer and generalization to novel ropes with unknown parameters. } \vspace{-3mm}
    \label{fig:challenge}
    \vspace{-3mm}
\end{figure*}

\begin{itemize}[leftmargin=3mm]
\item \textbf{Iterative action refinement}: Instead of directly inferring the optimal action from the goal, \names~starts with a best guess action and iteratively refines it to move toward the goal. For tasks with repeatable dynamics, this iterative approach allows the system to achieve high precision and be robust against errors from action, observation, and model prediction.

\item \textbf{Delta dynamics}: 
Within each iteration, \names~ learns to predict an updated trajectory from an observed trajectory with small action perturbations. Compared to modeling the entire dynamical system (action in, trajectory out), this ``delta dynamics'' (especially its general direction) is much easier to predict and generalize to new objects, thereby, enabling quick online adaption.

\end{itemize}

Our  primary contribution is the \namel~(\names), a general learning framework for goal-conditioned dynamic manipulation. 
Despite only trained using simulation data, \names~can be directly applied to real-world hardware. Furthermore, \names~demonstrates impressive generalization capability to unseen object instances, out-of-distribution rope parameters, unmodeled physical effects, and varying robot embodiments.
Our experiments show that \names~can achieve pixel level accuracy for a wide variety of goals in both simulation and real robot environments (1.8cm and 2.6cm respectively).

\section{Related work}
In this section, we will focus on summarizing relevant prior methods for deformable object manipulation. 

\textbf{Goal-conditioned manipulation} 
is particularly difficult for deformable objects due to their high degree-of-freedom and under-actuation \cite{7989247, 8575448,goal_transporter,ken_dense_descriptor,fabric_vsf_2020,abbeel_early}. Previous work studied tasks such as rope knot-tying \cite{ken_dense_descriptor} and fabric folding \cite{fabric_vsf_2020}. Early methods attempt to transfer demonstration trajectories to unseen configurations \cite{abbeel_early}. Recently, deep learning has been used extensively \cite{goal_transporter,ken_dense_descriptor,fabric_vsf_2020} to enable policy generalization to unseen objects or multiple tasks. However, these methods only consider quasi-static actions, where action effects are inferred mostly from object geometry alone. 
While quasi-static manipulations has shown to be sufficient to solve many tasks, the resulting systems are often slow and inefficient.

\textbf{Dynamic manipulation}
takes advantage of momentum (in addition to kinematic, static and quasi-static mechanisms) to increase load capacity or enable motions to have manipulands extend outside of the robot's nominal reach range \cite{dynamic_og,tossingbot,8344486,7139990,8280543,fast_hand,japan_shake}. Deformable objects pose a unique generalization challenge for dynamic manipulation policies due to complex physical properties and a particularly large sim2real gap.
Using reinforcement learning, Jangir et al. \cite{dyn_cloth_rl} proposes a system to dynamically fold and place cloth in various goal configurations. However, thousands of interactions are required for training. Similarly, Lim et al.\cite{casting} closes the sim2real gap by algorithmically tuning physics simulation parameters with real-world observations. But this process needs to be repeated for every unseen object instance. 
Leveraging visual feedback and domain randomization, Hietala et al \cite{finnish} is able to successfully transfer a manipulation policy trained in simulation to a real robotic setup. However, the policy is limited to a narrow range of goals such as folding clothes exactly in half. In comparison, our method is able to generalize to both unseen real-world object instances as well as wide variety of goals, making it a step closer to real-world applications.

\textbf{Trajectory optimization} uses analytical models \cite{fast_hand} or numerical simulation \cite{peter_model,bio_whip,optimal_planning,trajopt_berkeley,japan_rope} to generate mostly open-loop solutions for deformable object manipulation. Once a model has been designed, or automatically identified\cite{japan_rope}, the optimization problem can be solved using methods such as direct collocation \cite{trajopt_berkeley}, single shooting \cite{peter_model} or black-box optimization \cite{bio_whip}. These methods are generally capable of handling a wide range of goals and generalizing to many object instances, as long as a model is available for each new object. However, direct execution of optimal trajectory solutions on hardware (i.e., the OptSim method described later) does not work well for our tasks due to the large sim2real gap. In contrast, our method bridges this gap using feedback from previous trajectories without an explicit dynamics model.

\textbf{Iterative Learning Control (ILC)} leverages feedback from previous iterations to improve the accuracy of a repeatable system \cite{ilc_book, ilc_survey}. Most ILC algorithms tackle the task of trajectory tracking control, compensating for repeated disturbances against a reference trajectory. 
However, when the objective function is non-convex and difficult to optimize (like the two tasks we investigate in this paper), the reference trajectory is hard to obtain. Hence, ILC can not be directly applied.
In addition, existing ILC algorithms often assumes known control direction or Jacobian which is difficult to obtain for our task. In contrast, our method is able to iteratively solve these tasks despite its complexity.

\section{Problem setup}
\label{sec:setup}
We use two domains as examples of general dynamic manipulation tasks, one featuring 1-D deformable objects (ropes) and the other focusing on 2-D deformable objects (table cloths). For both tasks, the physical parameters of the objects are not known to the algorithm a priori.
During testing, a single policy is used for each task and evaluated across a diverse set of manipulands and robot embodiments, demonstrating our method's strong generalization capability to objects significantly outside of its training distribution.

\textbf{Rope whipping.}
The task is to hit a target location in the air with the tip of a rope attached to a UR5 robot (Fig \ref{fig:teaser} first row). The range of target locations far exceed the robot's reach range, requiring the robot to swing the rope dynamically. To reach sufficient speed in practice, we extended the UR5's end-effector with a 50cm long wooden stick. We use a parametric action primitive to describe the robot's movement. The action space for the whipping primitive is $a = (v, J_2, J_3)$, where $J_2 \in [90,-30]^{\circ}$ and $J_3 \in [-110,-290]^{\circ}$ are the target angles for joints two and three, respectively, and $v \in [1.0,3.14]$ \si{rad/s} is the maximum permissible angular velocity across all joints.

This action primitive considers only 2D movement in the Y-Z plane, where target locations are defined in-plane --- it is sufficient for this task, since out-of-plane goals could be reached simply by rotating the robot's base joint. 
The trajectory of the rope tip $T_i$ is tracked and rasterized to a $256^2$ image as observation input.
The distance metric $\mathcal{D}(T_i,g)$ for this task is defined as the minimum distance from any point on the trajectory $T_i$ to the goal location $g$.

\begin{figure*}[t]
    \includegraphics[width=\linewidth]{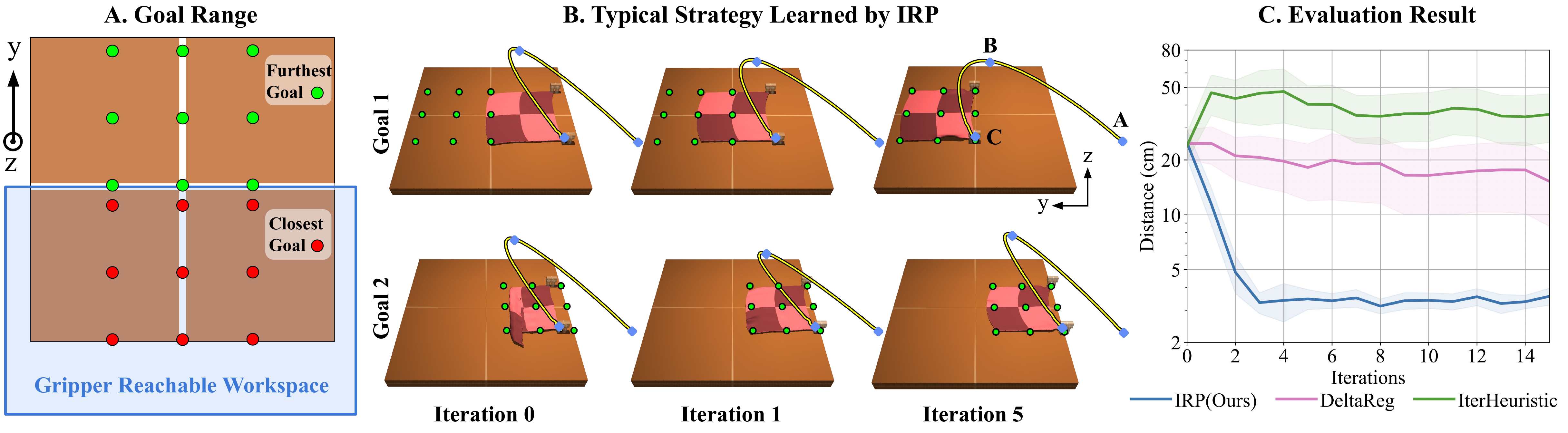}
    \caption{\textbf{Cloth placing task} A) Some goals are placed outside the arm's reach range, therefore requiring dynamic actions. 
    B) Top row: our algorithm adjusts the trajectory to reach a far-away goal. Bottom row: our algorithm reduces action velocity to eliminate undesired folding.
    C) Mean distance to goal keypoints averages across all test cloth and goals, shown with a 95\% confidence interval. Our method yields significantly better performance compared to the [deltaReg] baseline. Note that due to the stretching of the cloth and thickness of collision capsules, it is not possible to achieve 0 error. }
    \label{fig:cloth_plot} \vspace{-4mm}
\end{figure*}

\textbf{Cloth placement.}
The task is to place a square cloth on the table that reaches the goal configuration specified by 9 keypoints on the cloth (shown in Fig. \ref{fig:teaser} bottom row). Again, the goal configurations are further than the arm's maximum reach range, requiring dynamic actions. We assume that the cloth is gripped by two grippers that move in sync and consider the case where the grippers only move within the Y-Z plane, since out-of-plane goals can be reached by horizontally translating the trajectory. Gripper trajectory is parameterized as a cubic spline with $3$ via-points, evenly spaced temporally, in the Y-Z plane, Fig. \ref{fig:cloth_plot}. The start point and the Z coordinate for the end point are fixed. The remaining 3 degrees of freedom --- Y,Z locations for the second via-point and the Y locations of the third via-point--- as well as the total duration of the action, constitute the $4$ dimensional action space. The trajectory for all $9$ keypoints on the cloth $T_i$ is used as input observation to the algorithm. 
The distance metric $\mathcal{D}(T_i,g)$ for this task is defined as the mean keypoint distance between the cloth's final configuration and target configuration. 

\section{Approach}

\begin{figure}
    \centering
    \includegraphics[width=\linewidth]{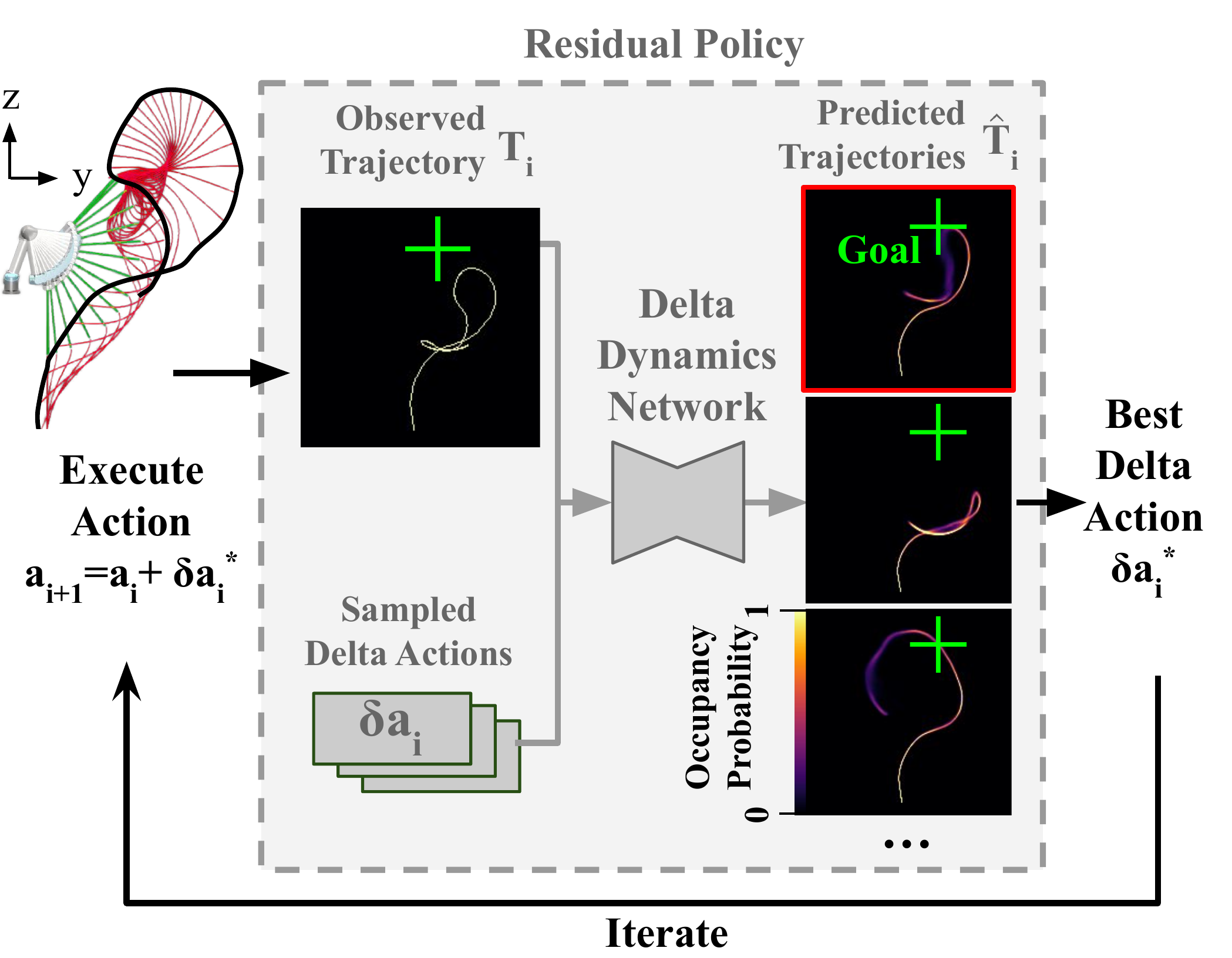}
    \caption{\textbf{\namel}. During each action execution, the tip trajectory is recorded as an image. For each sampled delta action $\delta a$, the Delta Dynamics Network predicts trajectories resulting from applying each delta action. The delta action that is predicted to minimize distance to the target is selected for execution during the next iteration.}
    \label{fig:method} \vspace{-1mm}
\end{figure}

\begin{algorithm}[t]
\caption{Iterative Residual Policy} \label{alg:IRP}
\begin{algorithmic}[1] 
\STATE Input: Goal position $g$
\STATE Initialize action $a_0 \in \mathbb{R}^{N_a}$ \hspace{0.1in} $\triangleright$ \S\ref{sec:method_sample} Initial action
\WHILE{$i <$ max\_step}
    \STATE $T_i$ = robot\_execution($a_i$) 
    \STATE $d_i$ = $\mathcal{D}$($T_i$, $g$) \hspace{0.5in} $\triangleright$ \S\ref{sec:setup} Defined for each task
    \IF{$d_i < d_{stop}$ }
        \STATE break;
    \ENDIF
    \STATE ${\delta a_i^{0:N_s}}$ = sample\_action($d_i$)  \hspace{0.2in} $\triangleright$ \S\ref{sec:method_sample} Delta action
    \STATE ${\hat{T}_i^{0:N_s}}$ = delta\_dynamics($\delta a_i^{0:N_s}$, $T_i$)   {\hspace{0.2in} $\triangleright$ \S\ref{sec:method_RDM}}
    \STATE $j^* = \argmin_{j}$($\mathcal{D}$($\hat{T}_i^{j}$,$g$))
    \STATE $a_{i+1}$ = $a_{i} + \delta a_i^{j^*}$
\ENDWHILE
\end{algorithmic}
\end{algorithm}

When attempting to hit a target with an unfamiliar rope, humans will generally not succeed on their first try. However, using physical intuition, people can usually \textbf{predict the effect of adjusting their actions}; when swinging a rope, for example, larger force will make the rope tip swing higher. Although the prediction is not perfect, people often adjust their actions in the right direction and quickly drive down their error after a few interactions.

Building on this intuition, we propose \textbf{Iterative Residual Policy} for goal-conditioned dynamic manipulation of deformable objects. Given an observation of trajectory $T_i$ by action $a_i$, we sample a set of potential delta actions $\delta a_i$ and predict the resulting trajectory $\hat{T}_i$ for each delta action. We then evaluate each of the predicted trajectories based on their minimal distance to the goal $d_i$, and greedily select the optimal delta action $\delta a_i^{j*}$ to execute in the next iteration. Trained using simulation data alone, this method is able to achieve high accuracy and generalize to out-of-distribution objects on our real-world robot setup. The following sections provide details for the key algorithm components and design decisions. 

\subsection{Delta Dynamics Network}
\label{sec:method_RDM}
The delta dynamics network takes in an observed trajectory $T_i$ and a delta action $\delta a_i^j$ as input and predicts the trajectory ${\hat{T}_i^{j}}$ after the delta action is applied. This network is used at every iteration for action optimization.

\textbf{Trajectory representation.}
To spatially represent the trajectory $T_i$, we rasterize the it into a $256\times256$ image projected onto the Y-Z plane. $T_i$ covers $\pm 3m$ area from the robot base joint. 
The pixel values correspond to occupancy probability; the observed trajectories have a binary pixel and the predicted trajectories are real-valued: $p \in [0,1]$. For the cloth placing task, the trajectory image of all 9 keypoints are stacked channel-wise.
Due to small differences in initial condition, hardware precision, ambient airflow as well as other disturbances, the trajectory is not always the same for each action. Moreover, different segments of the trajectory exhibit different sensitivity to noise. 

\textbf{Action representation.}
We broadcast the $N_a$ dimensional delta action $\delta a_i$ into a $N_a$ channel image and then concatenate them with the trajectory images before feeding them into the network. This spatially aligned representation allows us to use a standard CNN  architecture designed for image segmentation while ensuring the action information is available in the receptive field of every neuron.

\textbf{Network and loss.}
We use DeepLabV3+ \cite{deeplabv3} for our network architecture. During training, we uniformly select actions as $a_i$ and sample $\delta a_i$ with a Gaussian distribution (SD=0.125). The resulting trajectories in simulation are used as supervision. The network is trained with Binary Cross Entropy Loss and the AdamW optimizer \cite{adamw} with learning rate of 0.001 and weight decay of $1\times 10^{-6}$.

\subsection{Interactive Action Sampling and Selection}
\label{sec:method_sample}
\textbf{Initial action.} In the rope whipping task, the initial action $a_0$ is selected using the best average action for all training rope given a specific goal. This action can be computed efficiently with our offline training dataset.
The same action is used for a goal regardless of the rope parameters.
In the cloth placement task, a constant initial action is used for all goals to ensure all keypoint trajectories are observable in the first iteration.

\textbf{Delta action.}
Since each dimension of the action space has been scaled to between 0 and 1, we sample $N_s=128$ delta actions $\delta a_i$ for each iteration from a spherical Gaussian distribution with 0 mean and standard deviation $\sigma$. To reduce overshooting and accelerate convergence, we select $\sigma = 0.5 \times d_i$, where $d_i =\mathcal{D}(T_i,g)$.

\textbf{Action selection.}
After the network predicts the raw trajectories image for all delta action sample $\delta a_i^{0:N}$, we threshold the prediction with $t=0.2$, creating a set of binary trajectory image $\hat{T}_i^{0:N}$. 
Then, the delta action $\delta a_i^{j*}$ associated with smallest distance $\mathcal{D}(\hat{T}_i^{j},g)$ is selected to compute next action: $a_{i+1} = a_i + \delta a_i^j$. 
In real-world experiments, the optimization stops when the policy reaches $\mathrm{d\_stop}=2cm$ error or maximum iteration $\mathrm{max\_step}=10$ is reached. In simulation, the optimization executes for exactly $16$ iterations. The algorithm is summarized as pseudocode in Algorithm \ref{alg:IRP}.

\begin{figure*}[t]
    \centering
    \includegraphics[width=\linewidth]{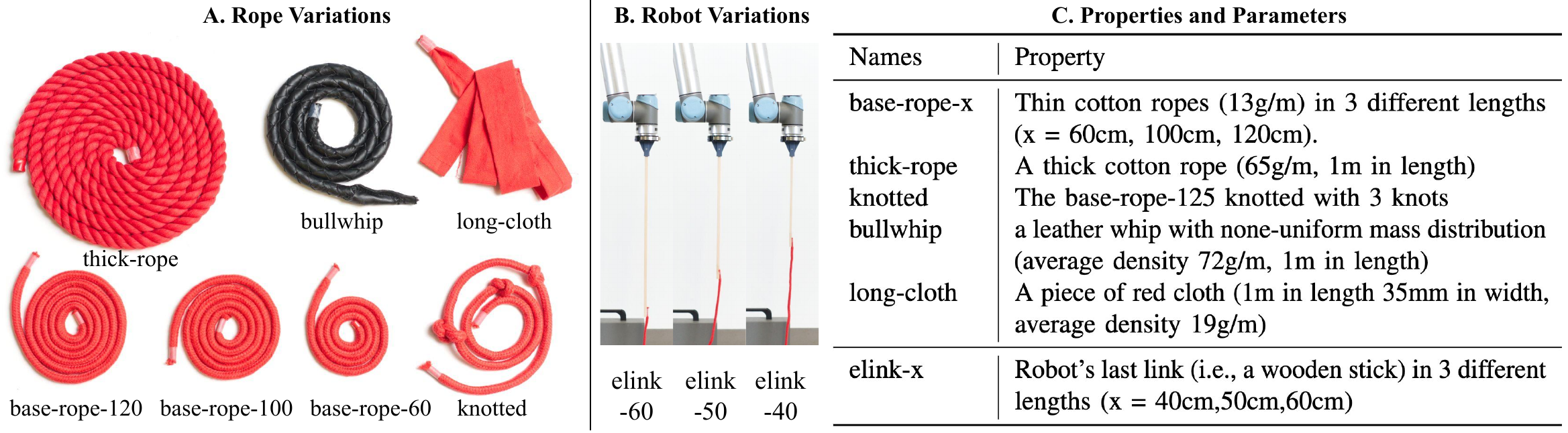}
    \caption{ \label{fig:testcase} \textbf{Real-world test cases.} A) Real-world testing ropes used to test our algorithm's generalization to untrained material, length, mass distribution and aerodynamics. B) By changing the the length of last link, the mapping from action parameters to effective swing speed and end-effector trajectory are changed simultaneously.  C) Table for key properties and parameters for each real world test case.}   
    \vspace{-3mm}
\end{figure*}

\subsection{Training Data Generation}
\label{sec:data_generation}

\begin{figure*}[t]
    \centering
    \includegraphics[width=0.264\linewidth]{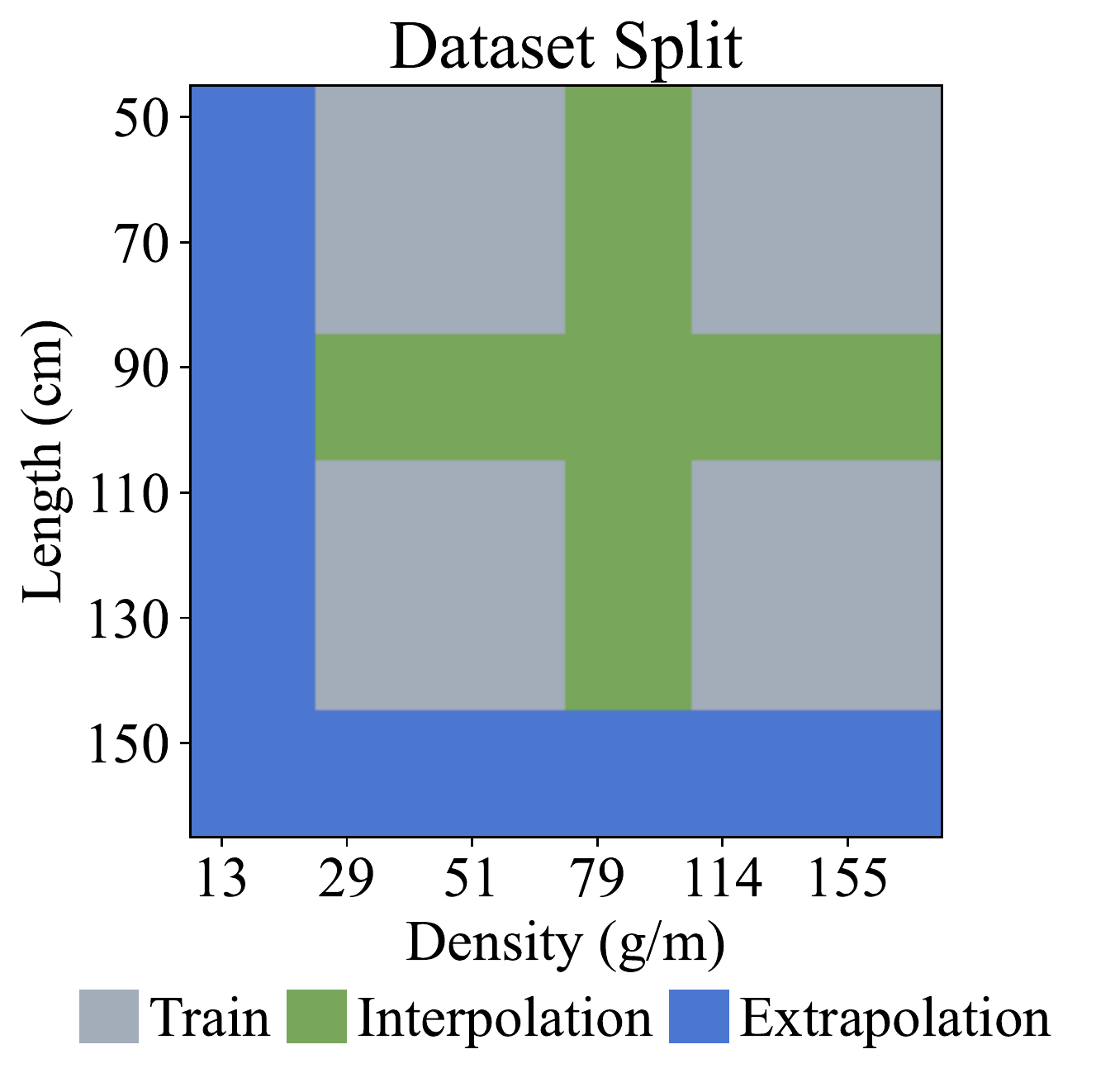}
    \hspace{-4mm}
    \includegraphics[width=0.736\linewidth]{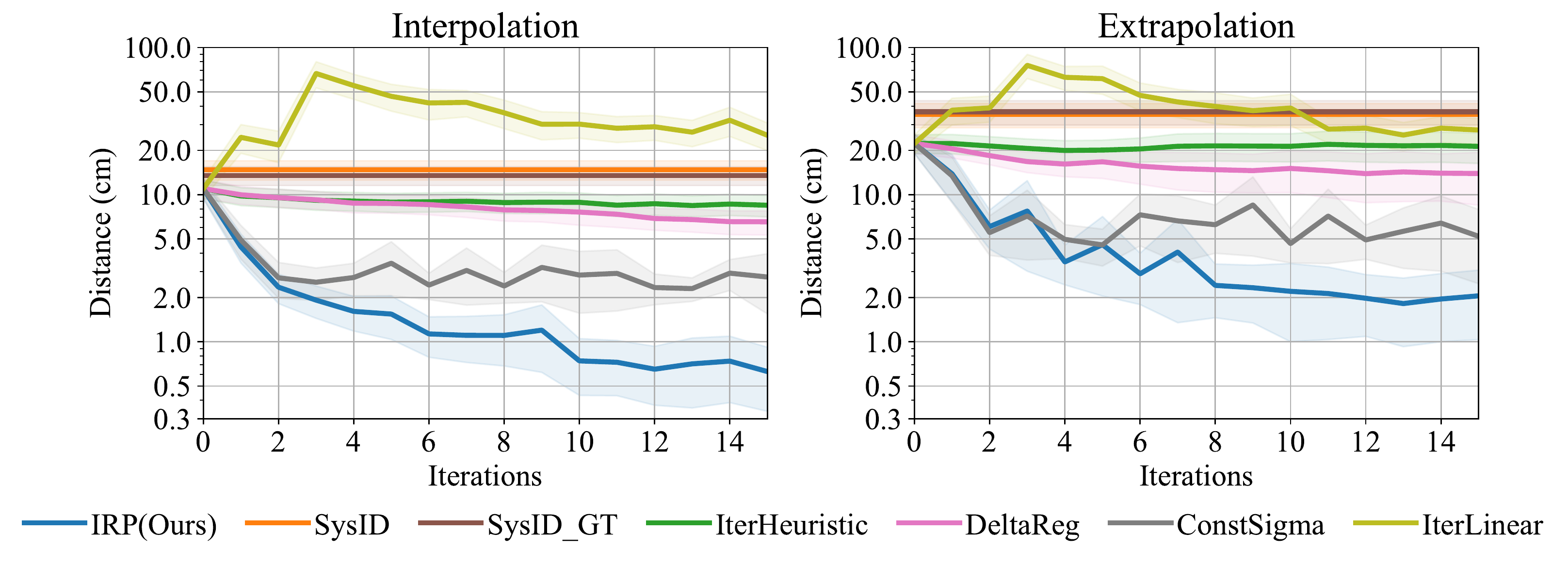}
    \vspace{-1mm}
    \caption{\textbf{Rope Experiments in Simulation.} Left: $12\times12$ grid of rope parameters used in simulation experiment (grey: training, orange: interpolation, blue: extrapolation). Middle and Right:  Distance to goal in cm averaged across 5 ropes and 25 goals for each method, shown with 95\% confidence interval. The result of single-step methods (SySID, SySID\_GT) are shown as lines for visualization.
    Testing rope with parameters that extrapolate  beyond training ropes parameters (middle) typically result in higher distance errors for most of the alternative approaches. However, \names is able to achieve low distance errors for both scenarios (0.4cm and 1.5cm respectively).} 
    \vspace{-3mm} 
    \label{fig:sim}
\end{figure*}

\textbf{Rope whipping.}
For training and validation, we built a simulation environment in MuJoCo \cite{mujoco}. The rope is simulated using 25 linked capsules and to generate different ropes, varying two parameters: length and density. The range of both parameters as well as the training/testing split is shown in Fig. \ref{fig:sim} left.  
The rope parameters are sampled from a $12^2$ grid. For each rope, we simulate $50^3$ actions within the set box constraint for speed $[1,3.14]$ \si{ rad/s}, joint 2 $ \in [90, -30]^{\circ}$ and joint 3 $\in [-110, -290]^{\circ}$. In addition, each action was repeated 3 times with different random perturbations on the initial state to capture the stochastic nature of the trajectory. The entire dataset contains 54 million trajectories.

Clearly this model does not capture all the variations in real-world ropes, and the simulation itself presents different dynamics from the real world due to unmodeled factors such as aerodynamics, collision with the floor, etc. Despite the significant sim-to-real gap, we will later show that our model trained on these simulated ropes generalizes very well in the real world with significantly out-of-distribution ropes. 

\textbf{Cloth placement.}
Similarly, the training data for cloth is collected from a MuJoCo simulation environment. The square cloth is simulated as a $13^2$ grid of capsules, skinned as a cloth for visualization. We again vary 2 parameters: size (ranged between 0.4 to 0.6 m) and density (ranged between 0.2 to 1.4 \si{kg/m^2}). For each cloth parameter, we sample speed and 3 via-point parameters from a $8\times16^3$ grid. In total, the dataset contains 131 thousand examples.

\subsection{Real world system setup.}
We conducted real-world experiments for the rope whipping task with an UR5-CB3 robot augmented with a wooden extension. We used a single RGB camera (Stereolab ZED 2i) running at 720p 60fps for tracking the tip of the rope. Due to the limited resolution and large motion range, stereo tracking does not provide sufficient depth accuracy and reliability. We therefore assume that the rope tip only moves in a 2D plane, allowing us to reduce the perception problem to that of 2D tracking. We placed the camera 2.4m away from the robot, and calibrated the homography between the image plane and the robot Y-Z plane with manually labeled point correspondences. We trained a keypoint detection model based on DeepLabV3+ with 400 hand labeled images.
Note that we use a higher image resolution (720p instead of 256p) for tracking results with less than 1.3cm tracking error in the validation set.

\section{Evaluation}
In our experiments we seek to validate the system's generalization capability to 1) novel object physical parameters 2) real-world dynamics, and 3) robot hardware embodiment. In all following experiments, both in simulation and the real-world, we use the same model trained with simulation training ropes generated in Sec. \ref{sec:data_generation}. The cloth placing task is evaluated in simulation only and is discussed in Sec. \ref{sec:cloth_eval}

\begin{figure}[t]
    \centering
    \includegraphics[width=\linewidth]{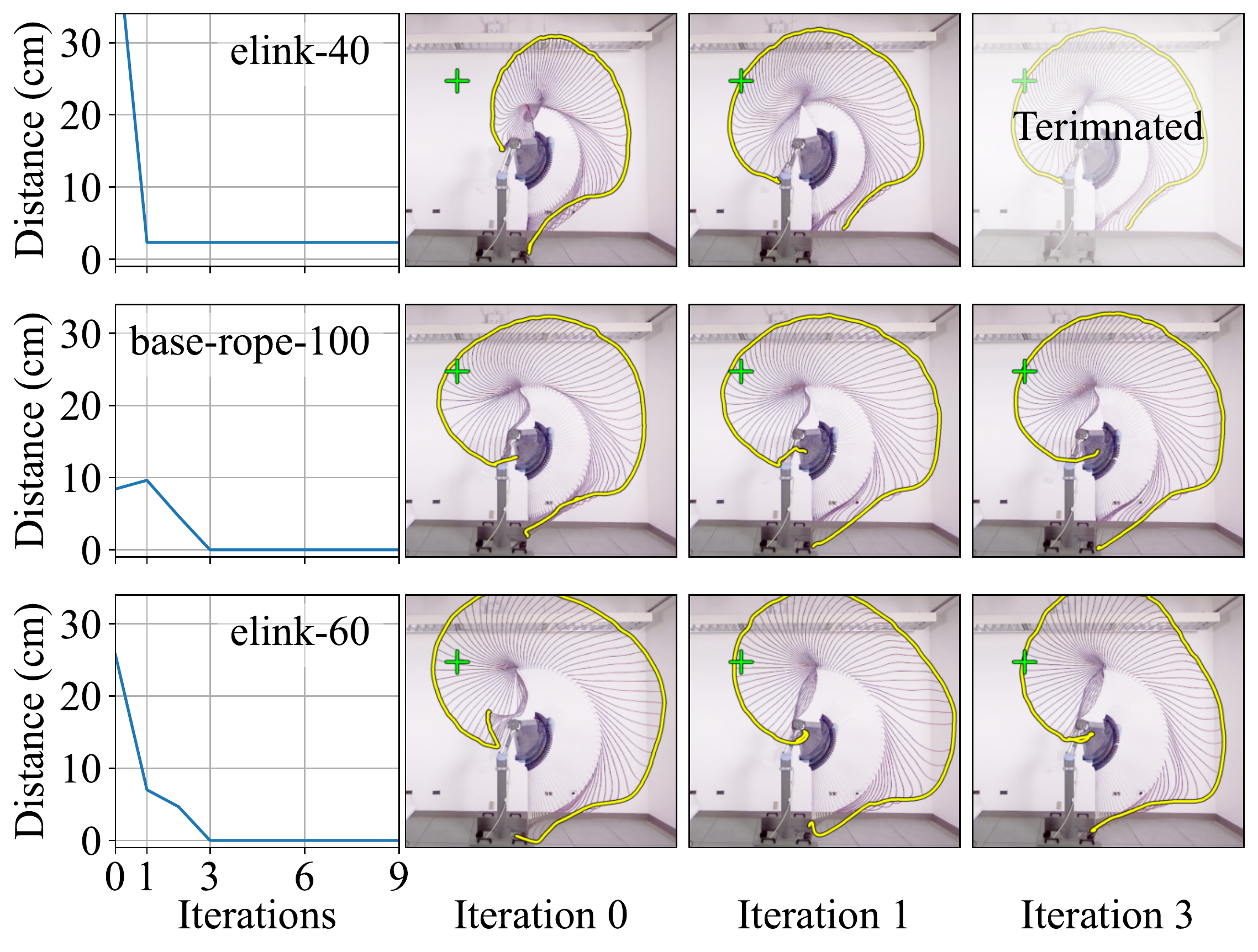}
    \caption{\textbf{Adaptation to robot embodiment.}  We varies the robot's last link with 3 different lengths (40,50,60cm), which effectively changes the mapping between the actions and their effects. As expected, the robot with a different link than training (elink-40/-60) has a higher initial error. However, for all variations, the system is able to reach the goal by adjusting its action according to the visual feedback.} 
    \label{fig:real_stick} 
    \vspace{-6mm}
\end{figure}

\subsection{Rope Whipping Task}
\textbf{Metrics.}
We measure the performance of these algorithms using the minimum distance to the goal (in cm) at each step. The faster that an algorithm reaches a certain distance the better. We allow the maximum iteration to be 16 in simulation and 10 in real-world experiments. Note that the pixel size is around 2.3cm in our trajectory image encoding and the real-world tracking accuracy is around 1.3cm.

\textbf{Test Cases.}
To test the system's ability to generalize to different rope parameters, we used the following testing ropes: 
\begin{itemize}[leftmargin=*]
    \item Simulated ropes with parameters in the \textbf{interpolation} regime relative to the training rope parameters (Fig. \ref{fig:sim} green).
    \item Simulated ropes with parameters in the \textbf{extrapolation} regime relative to the training rope parameters (Fig. \ref{fig:sim} blue).
    \item \textbf{Real-world ropes} with different material, length, and mass distribution (Fig. \ref{fig:testcase} C). We modeled the simulation rope after the [base-rope], therefore all other ropes  are significantly out-of-distribution. In particular, the [bullwhip], and [knotted-rope] have a non-uniform mass distribution and the [long-cloth] has a larger surface area to density ratio, hence, experiences much larger aerodynamic effects (unmodeled in simulation) than to other ropes.     
\end{itemize}

\textbf{Robot hardware embodiment.} We validate the system's generalization capability to different robot hardware embodiments by sampling the robot's last link length from 3 different lengths, [elink-x] where x = 40cm,50cm,60cm. By changing robot's last link, we are effectively changing the mapping between the robot's action parameter and its physical embodiment, which require the system to adapt according to the new trajectory observations. Fig. \ref{fig:real_stick} shows the example results for this experiment, where \names~ converges to 3 different actions for the same goal.

\begin{table}[t]
\centering
    \setlength\tabcolsep{4pt}    
    \begin{tabular}{l|rrrrrr}
    \toprule
    {} &  OptSim &  AVG &  il-3 &  il-9 &  IRP-3 &  IRP-9 \\
    rope          &         &      &       &       &        &        \\
    \midrule
    base-rope-100 &    21.6 & 15.6 &  31.5 &  13.4 &    3.2 &    \textbf{1.3} \\
    base-rope-120 &    14.4 & 16.5 &  51.9 &  22.5 &    \textbf{1.9} &    \textbf{1.9} \\
    base-rope-60  &    14.5 & 19.9 &  28.4 &  28.0 &    8.9 &    \textbf{5.5} \\
    knotted       &    14.2 &  8.3 &  17.1 &   9.2 &    \textbf{2.6} &    \textbf{2.6} \\
    thick-rope    &    11.9 & 19.7 &  29.2 &  10.7 &   11.7 &    \textbf{3.2} \\
    long-cloth    &    15.8 & 59.6 &  72.2 &  16.8 &   14.0 &    \textbf{1.9} \\
    bullwhip      &    17.0 & 28.7 &  50.5 &   8.4 &    9.0 &    \textbf{1.9} \\
    \midrule
    elink-40      &    16.0 & 28.4 &  77.5 &  29.3 &   13.3 &    \textbf{6.1} \\
    elink-60      &    11.9 & 14.6 &  30.5 &  18.5 &    5.4 &    \textbf{3.8} \\
    \bottomrule
    \end{tabular}

    \caption{\textbf{Realworld Evaluation.} Distance to goal in cm for 7 unseen real world ropes and 2 robot hardware embodiments. [long-cloth] and [bullwhip] are particularly out-of-distribution. Our method significantly outperforms the other approaches we evaluated, demonstrating surprisingly strong generalization to unseen rope parameters, action definitions and simulation/reality divergence. IL: iterLinear. X-3, X-9 error measured in iteration 3 and 9.} \vspace{-4mm}
    \label{tab:real_eval} 
\end{table}

\textbf{Algorithm comparisons.} 
To validate the major design decisions and their impacts on performance, we compare with following alternative approaches: 
\begin{itemize}[leftmargin=3mm]
    \item \textbf{System identification (SysID, SysID\_GT)}:  This method first identifies key system parameters and then infers the action using these parameters. We give this baseline a head start by providing the true varying parameters in simulation: length and density. To ensure fairness, we use 16 fixed system identification actions to collect observations. Another 4-layer MLP is trained to regress the optimal action from estimated rope parameters. [SysID\_GT] is a variant of this baseline, where we assume the ground truth rope parameters are given.
    
    \item \textbf{Iterative control with heuristic (iterHeuristic):} A heuristic strategy that increases both the speed and amplitude of the action if the trajectory intersects with the line segment between the goal and the origin (the goal is outside the trajectory) and decreases otherwise. If the trajectory does not intersect with the ray from the origin to the goal, the algorithm terminates to prevent increasing error.
    
    \item \textbf{Iterative control with linear model [iterLinear]}: 
    Instead of leveraging the residual dynamics model learned from offline data, a linear model of the plant is fitted on the trajectories observed so far, updated at each step. The action is adjusted at each step by minimizing the shortest distance to the goal using the linear model.
    
    \item \textbf{Direct delta action regression [DeltaReg].} Instead of using sampled delta actions as input, this method directly regresses the optimal delta action for the goal and the observed trajectory. The model is trained with MSE loss.  We test it with the same number of iterations.
    
    \item \textbf{No adaptive action sampling [ConstSigma].} An ablation of our method which uses a fixed sigma for action sampling.
    
    \item \textbf{Average action in simulation (AVG):} The action that minimizes the average distance to goal in the training set, regardless of rope parameters, this is also the action we used to initialize our method at step one. 
    
    \item \textbf{Optimal control in simulation (OptSim):} This method estimates optimal actions in simulation with measured rope parameters from a real-world (i.e., length, density and width). This method represents an oracle for model-based control using our simulator. It is used only in real-world experiments to quantify the sim2real gap.
\end{itemize}

\begin{figure}[t]
    \centering
    \includegraphics[width=\linewidth]{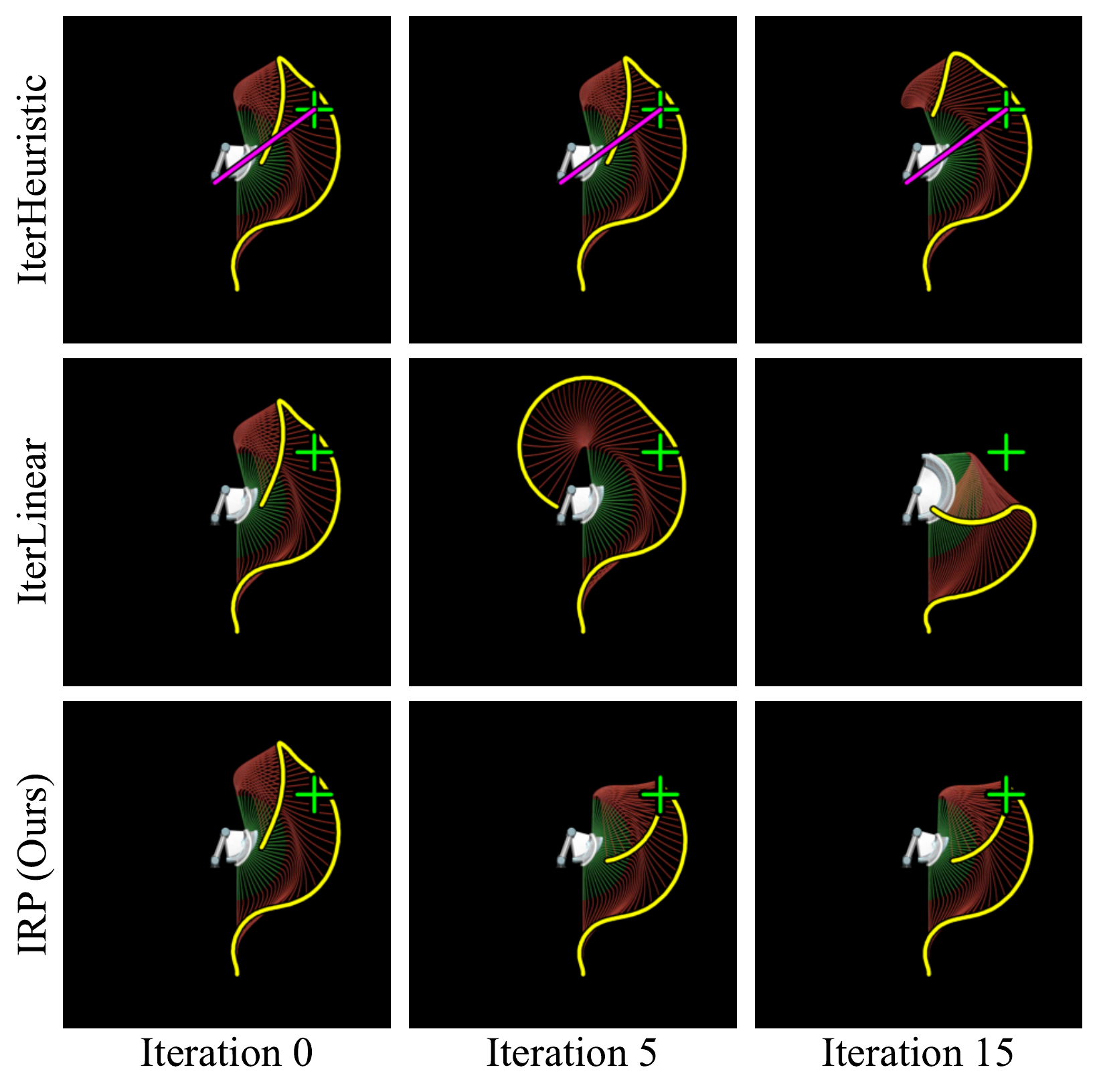}
    \caption{\textbf{Comparison.} In this example, the rope trajectory switches between several discrete modes. As a result, IterHeuristic gets stuck in a sub-optimal orbit and iterLinear failed to approximate the switching behavior. In contrast, our method accurately predicts the switching behavior and reaches the goal.} \vspace{-4mm}
    \label{fig:comparison}
    \vspace{-3mm}
\end{figure}

\subsection{Experimental Analysis}
\textbf{Benefits of learning residual dynamics.} 

We hypothesize that this formulation eases the learning process and maintains the flexibility of representing complex non-linear dynamics. To validate this hypothesis, we compared to other alternatives for modeling the system.



Compared to the system identification method [SysID, SySId\_GT], \names~can better handle unmodeled physical parameters presented in real-world ropes (e.g., aerodynamics or stiffness), which is captured in the input trajectory but not in the predefined system parameters (e.g., length and density). As a result, the algorithm can adapt to real-world ropes much better than SysID methods. 

Compared to [DeltaReg], \names~ better models the multi-modal aspect of the solution space.
Oftentimes there are multiple actions could reach the same goal, where [DeltaReg] tends to predict the mean of all valid solution (due to its MSE objective), which is likely to be an invalid solution. 



Compared to [IterHeuristic], our learned \names~model can capture complexity in the rope dynamics, such as mode switching when crossing key velocity thresholds, that the heuristic approach cannot capture. The [IterHeuristic] policy essentially assumes that the tip of the rope swings in a full circle, with its radius controlled by the action's energy. However, when the energy is insufficient for the tip to swing past apex this assumption is incorrect, and therefore, lead to incorrect action prediction. 
Moreover, around the apex point, the rope trajectory quickly switches between several modes, causing [IterHeuristic] to get stuck in oscillating local minima (Fig. \ref{fig:comparison} first row).

Finally, compared to [iterLinear]: our \names~model is able to converge in fewer steps with lower distance. In many cases, the rope trajectories switch between several discrete modes for different actions in a highly non-linear fashion, which cannot be easily captured by the linear model (Fig. \ref{fig:comparison}  second row). In contrast, our model is able to learn a non-linear model from the diverse set of training trajectories.

\begin{figure}[t]
    \centering
    \includegraphics[width=\linewidth]{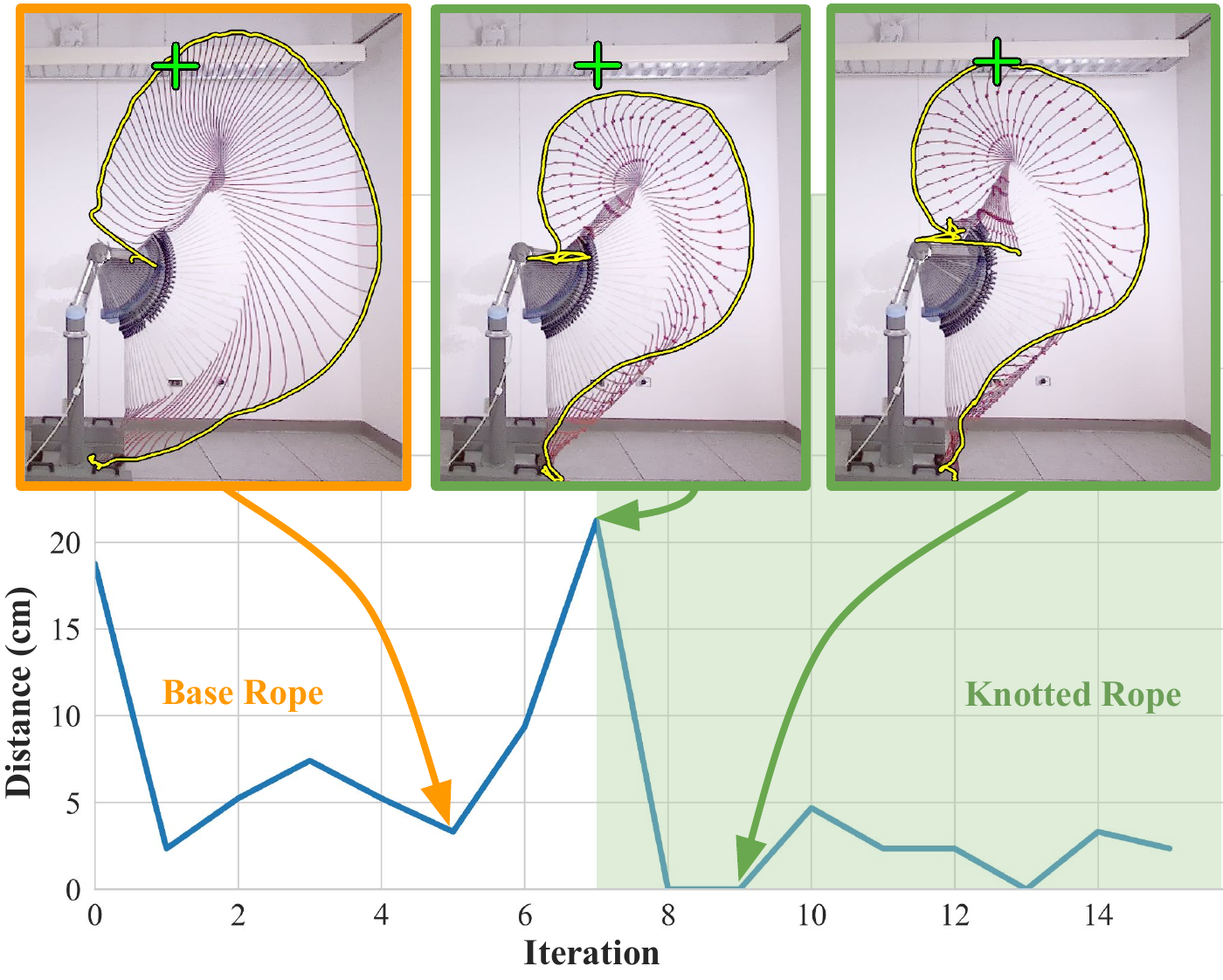}
    \caption{\textbf{Online adaptation to system  changes.} In this experiment, the robot first interacts with  [base-rope-100]. At step 6, the rope is knotted. We plot the distance to goal with respect to the steps.  While the tied knots significantly change the rope's length, density and mass distribution, the system is able to quickly adjust to the new system dynamics and achieve low errors.}
    \label{fig:online_adaptation} \vspace{-4mm}
\end{figure}
 
\begin{figure*}[t]
    \centering
    \includegraphics[width=\linewidth]{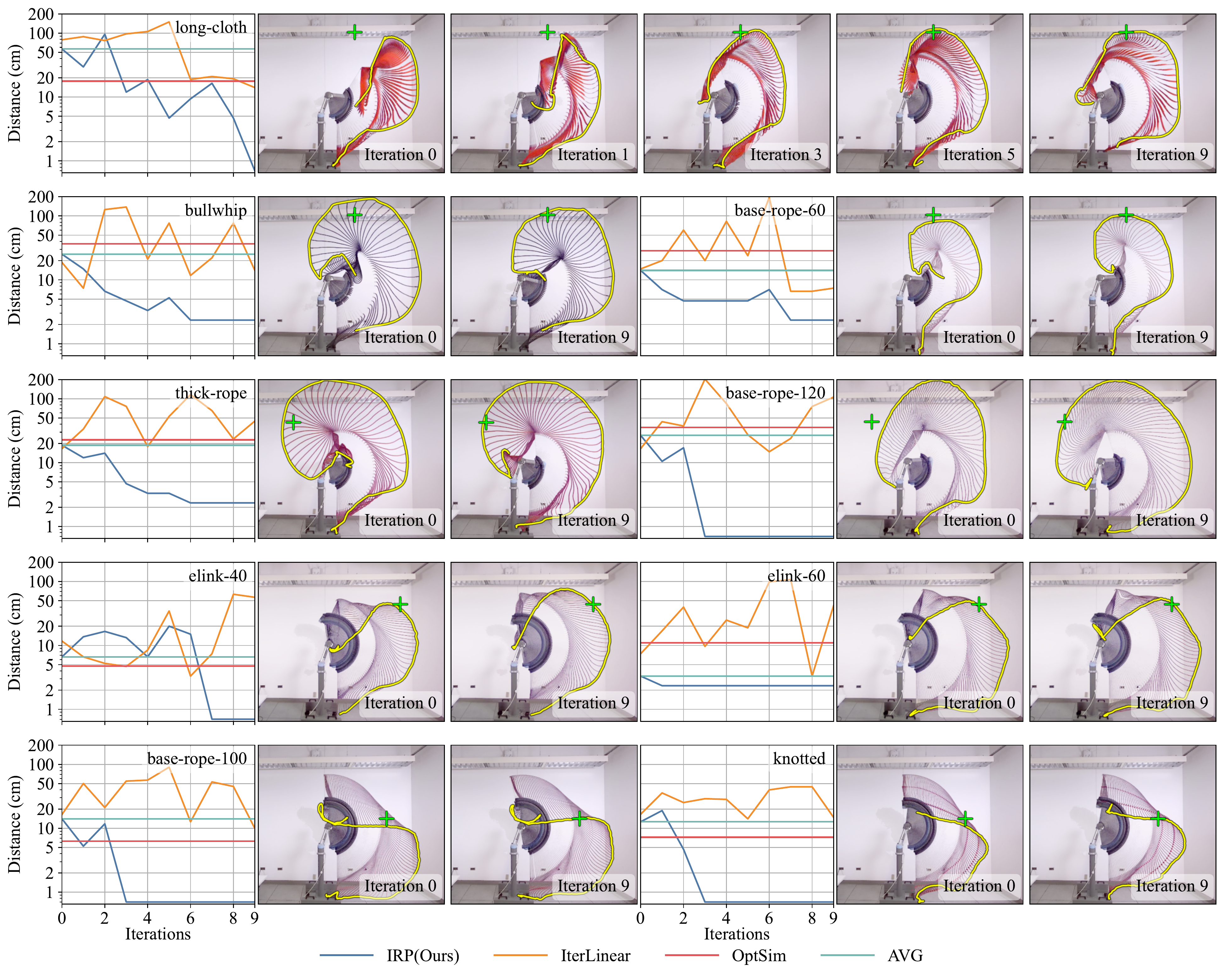} 
    \vspace{-6mm}
    \caption{\textbf{Real-world test results on different ropes.} Using the same network trained on simulation data only, our method is able to generalize to significantly out-of-distribution ropes and a wide range of goals. First row: step 0,1,3,5 and 9 for long-cloth test case. Second row: step 0, 9 for other test cases. OptSim is not able to achieve low error, indicating the large sim2real gap in this task. All distances are shown in \textbf{log scale}. iterLinear failed to find good solutions within 9 steps, since it uses online observations only and has limited model capacity. More results can be find in the supplemental video.} 
    \vspace{-4mm}
    \label{fig:real_detail}
\end{figure*}
\textbf{Benefits of iterative action refinements.}
With its iterative refinements, \names~can consistently improve its performance from the best average action.  On average, the error drops 94\% on interpolation experiments and 91\% drop on extrapolation experiments by using the additional trajectory observations. 
Moreover, the comparison with [ConstSigma] shows that the adaptive action sampling method (described in Sec. \ref{sec:method_sample}) can further improve the action samples and prevent overshooting around the optimal action --- demonstrated by higher error and variance for [ConstSigma] in Fig. \ref{fig:sim} C after step 4. 

\textbf{Comparison to optimal control in simulation.}
By computing the action using exhaustive search with manually measured parameters, the performance of [OptSim] represents an \textit{oracle} for trajectory optimization using our MuJoCo simulation environment, and it should achieve perfect result in the simulated environments.  
However, due to the instability of the dynamical system and unmodeled effects such as aerodynamics, [OptSim] still results in higher error compared to \names~in our real-world experiment (Fig \ref{fig:real_detail}, Tab. \ref{tab:real_eval}). 
This result demonstrates the importance of online adaptation for closing sim2real gaps, especially for complex dynamical systems such as the dynamic manipulation of deformable objects.

\textbf{Online adaptation to system changes.}
In this experiment, we stress test \names' robustness against unexpected system changes during the interaction steps. We first ask the robot to interact with the [base-rope-100]. After step 6, we tie four knots on the rope and observe the system behavior.  While the tied knots significantly changes the rope's length, density and mass distribution, we observe that the system is able to quickly adjust to the new system dynamics and regain good performance ( Fig. \ref{fig:online_adaptation}).

\section{Extension to Cloth Manipulation}
\label{sec:cloth_eval}
To demonstrate the generality of \names's, we applied the same method to a cloth placement task with minimal modifications.

\textbf{Test cases.}
We randomly sampled five unseen cloth parameter pairs (size, density) as our testing cases. For each cloth sample, we uniformly sampled 11 goal configurations from the range illustrated in Fig. \ref{fig:cloth_plot} A. Note that since test cloths all have different sizes, the specific target keypoint locations are adjusted accordingly. 

\textbf{Metrics.}
The performance is measured by the average distance to the goal (in cm) over all keypoints. Note that due to the stretching effects on the cloth and the thickness of MuJoCo's capsule model for collision, it is not possible to reach zero error for a target configuration with perfectly square and flat keypoint locations.

\textbf{Result.}
Fig \ref{fig:cloth_plot} B and C show the qualitative and quantitative results respectively. Fig \ref{fig:cloth_plot} B shows that \names~can adjust the action to different landing locations and shape of the cloth (e.g. to avoid folding). We compared with [DeltaReg] and [IterHeuristic], two of the best performing baseline methods from the rope whipping task. We modified the [IterHeuristic] to increase all action parameters if the landing keypoints are closer than the goal, and decreases otherwise, with the proportional gain set to $0.5$. On average, \names~achieves 3.5 cm error across 11 test goal configurations, yielding better performance and lower variance comparing to other methods.

\section{Conclusion}
This paper presents a general learning framework for goal-conditioned dynamic manipulation: \namel.  Our extensive experiments in both simulation and the real-world demonstrate its adaptability to many aspects of the system, including object parameters, real-world dynamics, and even robot hardware embodiment.
However, our method is not without limitations, one of which is the assumption of action repeatability, which can not be guaranteed in many applications.  We also assume full observability of the manipuland throughout the trajectory, which makes direct application of this approach difficult in highly cluttered scenarios.
Future work could explore using different visual feedback \cite{lin2021VCD,chi2021garmentnets} to deal with non-repeatable action scenarios and perception techniques leveraging temporal cues to handle partial observability \cite{xu2020learning}.


\section*{Acknowledgement} We would like to thank Zhenjia Xu, Huy Ha, Dale McConachie, Naveen Kuppuswamy for their helpful feedback and fruitful discussions. This work was supported by the Toyota Research Institute, NSF CMMI-2037101 and NSF IIS-2132519. We would like to thank Google for the UR5 robot hardware. The views and conclusions contained herein are those of the authors and should not be interpreted as necessarily representing the official policies, either expressed or implied, of the sponsors.

\bibliographystyle{plainnat}
\bibliography{references}

\end{document}